\documentclass[lettersize,journal]{IEEEtran}
\usepackage{amsmath,amsfonts}
\usepackage{algorithmic}
\usepackage{array}
\usepackage[caption=false,font=normalsize,labelfont=sf,textfont=sf]{subfig}
\usepackage{textcomp}
\usepackage{stfloats}
\usepackage{url}
\usepackage{verbatim}
\usepackage{graphicx}
\usepackage{cite}
\usepackage{booktabs}
\usepackage{multirow}
\usepackage{color}
\usepackage[table]{xcolor}

\usepackage{cleveref}
\newcommand{\etal}{\textit{et al}.}
\def\BibTeX{{\rm B\kern-.05em{\sc i\kern-.025em b}\kern-.08em
    T\kern-.1667em\lower.7ex\hbox{E}\kern-.125emX}}
\usepackage{balance}
\begin{document}
\title{OSM-Net: One-to-Many One-shot Talking Head Generation with Spontaneous Head Motions}
\author{Jin Liu, Xi Wang, Xiaomeng Fu, Yesheng Chai, Cai Yu, Jiao Dai, Jizhong Han \thanks{Jin Liu , Xiaomeng Fu and Cai Yu are with the Institute of Information Engineering, Chinese Academy of Sciences, Beijing, China, and also with the School of Cyber Security, University of Chinese Academy of Sciences, Beijing, China (e-mail: liujin@iie.ac.cn, fuxiaomeng@iie.ac.cn, caiyu@iie.ac.cn).

Xi Wang, Yesheng Chai, Jiao Dai and Jizhong Han are with the Institute of Information Engineering, Chinese Academy of Sciences, Beijing, China(w-mail: wangxi1@iie.ac.cn, chaiyesheng@iie.ac.cn, daijiao@iie.ac.cn, hanjizhong@iie.ac.cn).

}}

\markboth{}{}

\maketitle

\begin{abstract}
One-shot talking head generation has no explicit head movement reference, thus it is difficult to generate talking heads with head motions. 
Some existing works only edit the mouth area and generate still talking heads, leading to unreal talking head performance. Other works construct one-to-one mapping between audio signal and head motion sequences, introducing ambiguity correspondences into the mapping since people can behave differently in head motions when speaking the same content. This unreasonable mapping form fails to model the diversity and produces either nearly static or even exaggerated head motions, which are unnatural and strange. Therefore, the one-shot talking head generation task is actually a one-to-many ill-posed problem and people present diverse head motions when speaking. Based on the above observation, we propose OSM-Net, a \textit{one-to-many} one-shot talking head generation network with natural head motions. OSM-Net constructs a motion space that contains rich and various clip-level head motion features. Each basis of the space represents a feature of meaningful head motion in a clip rather than just a frame, thus providing more coherent and natural motion changes in talking heads. The driving audio is mapped into the motion space, around which various motion features can be sampled within a reasonable range to achieve the one-to-many mapping. Besides, the landmark constraint and time window feature input improve the accurate expression feature extraction and video generation.
Extensive experiments show that OSM-Net generates more natural realistic head motions under reasonable one-to-many mapping paradigm compared with other methods. 

\end{abstract}


\section{Introduction}
Given one source face image and one driving audio, one-shot talking head generation aims to synthesize a talking head video with reasonable facial animations corresponding to the driving audio ~\cite{zhang2021flow}. This task is of great significance to a wide range of multimedia applications, e.g. film making~\cite{prajwal2020lip,kim2019neural}, video dubbing ~\cite{kim2018deep}, digital human animation~\cite{zhou2018visemenet},  video conference~\cite{wang2021one}, video call compression~\cite{agarwal2022compressing} and fast short video creations~\cite{zhou2021pose}.  Due to its popularity, the talking head generation task has drawn great attention for a long time~\cite{chen2020comprises, mirsky2021creation,  liujin2022}.

A large number of one-shot talking head generation works~\cite{ song2018talking, zhu2018arbitrary,  zhou2019talking, chen2019hierarchical, kr2019towards, prajwal2020lip, vougioukas2020realistic} are proposed to only synchronize the audio signal and lip movements. They ignore to infer head motion information from driving audio and simply edit the mouth region, keeping the other areas of faces static. The still head pose is far from satisfactory for human observation since their performance is unnatural given the fixed facial contour and blending traces around the mouth area.

Hence, some methods~\cite{ zhou2021pose, wang2022progressive,liu2023opt} generate talking videos with head motions borrowed from another  auxiliary talking head video. The selection of additional video brings complexity and computational burden to the practical application.
Later, some works try to infer head pose sequences from input signals~\cite{zhou2020makelttalk , zhang2021flow, wang2021audio2head, wang2022one }. They mainly adopt LSTM-based modules and
perform one-to-one mapping between audio signals and talking head motion sequences. This mapping paradigm suffers from ambiguous correspondences between head motion and audio in the training dataset and fails to produce realistic head movements~\cite{wang2021audio2head}. For example, when speaking the same audio content, one subject may move towards the right while other subject to the left.  These examples will introduce ambiguity to the one-to-one mapping between audio signal and head motions. Thus the above methods generate talking heads with nearly static~\cite{zhou2020makelttalk, zhang2021flow} or exaggerated~\cite{wang2021audio2head} head motions.

Besides in audio-visual training dataset, as for real-world scenes, people could speak the same content with different natural head movements.  Assuming in the chorus scenario, it is weird for each singer to perform the same pattern of head motion changes. In metaverse, when different digital avatars are asked to speak fixed range of corpus, diverse head motions will increase the vitality of characters. Therefore, it is significant to build a one-to-many mapping in the one-shot talking head generation, i.e. producing diverse talking head videos with different realistic and natural head movements.

Based on the observation in datasets and real-world scenes, we believe that there exists a reasonable head motion space given a specific driving audio signal. In this way, talking head videos with diverse head motion sequences can be generated. However, there exist several challenges:  1) It is difficult to analyze the law of head motion changes and build up a meaningful head motion space. 2) The one-to-many mapping between audio signals to various head motion features can hardly be constructed. 3) The mouth shape accuracy and visual stability are hard to maintain during diverse motion changes since there exists a natural gap between audio-visual modality.

\begin{figure*}[t]
	\centering
	\includegraphics[width=\textwidth]{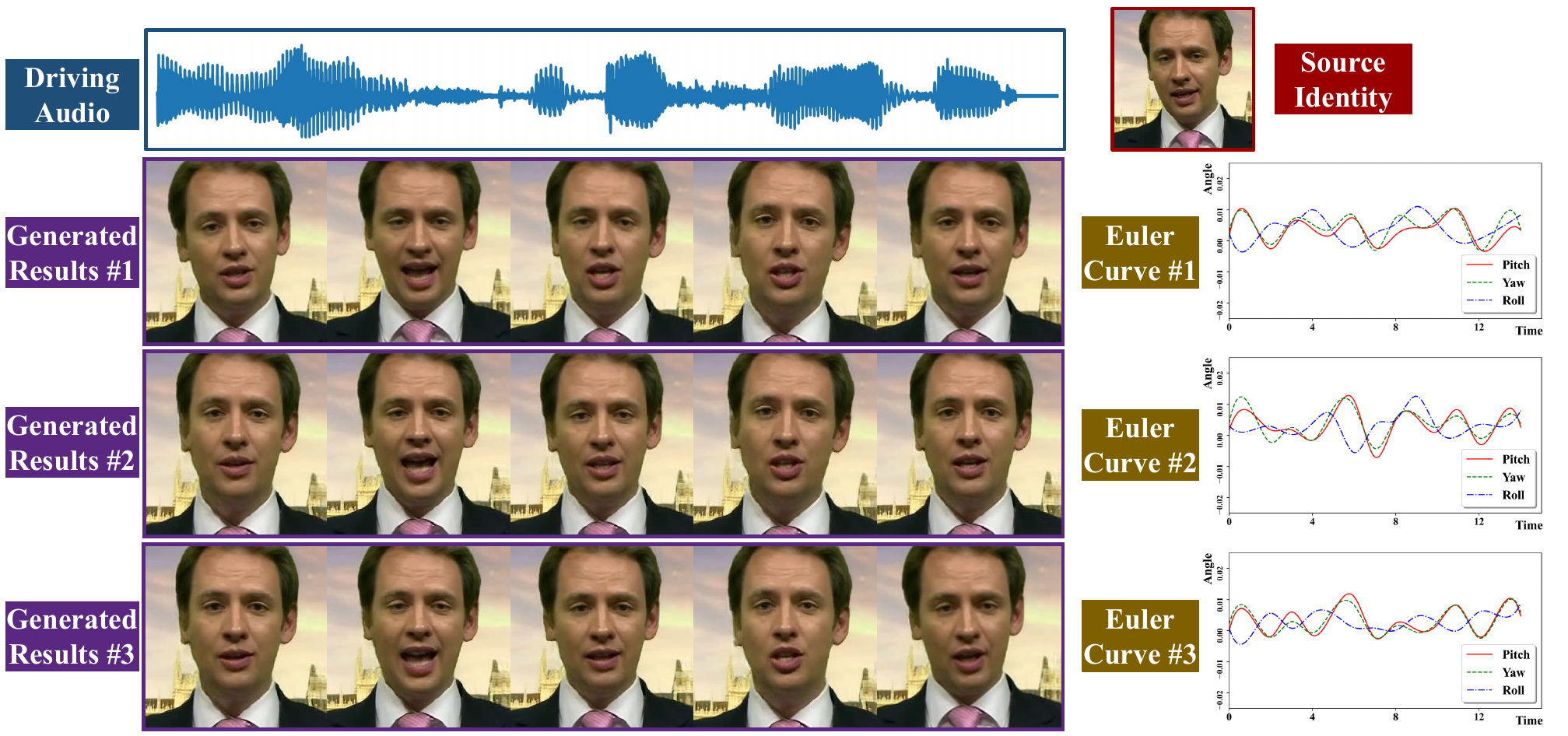} 
	\caption{Illustration of the one-to-many mapping of OSM-Net. Our method takes one source face as identity guidance and generates multiple talking head videos with diverse natural  head motions. All generated videos maintain accurate mouth shape, corresponding identity information and reasonable head motions. The corresponding euler angle changing curves through time dimension are shown, indicating the variation in roll-pitch-yaw angles of head movement.} 
	\label{fig:teaser}
\end{figure*}

In this work, to generate different talking head videos with natural diverse spontaneous head motions, we propose the one-to-many framework called OSM-Net, whose illustration is shown in Fig. \ref{fig:teaser}. Given one source identity image and driving audio signal, OSM-Net generates talking head videos with diverse spontaneous head motions.

Specifically, OSM-Net consists of three parts, i.e. Audio-Motion Mapping Network, Expression Feature Extractor, and Video Generator. 
The \emph{Audio-Motion Mapping Network} builds a one-to-many mapping between audio signal to head motion sequences. It first constructs a universal and meaningful clip-level motion space given learned head motion basis. Each basis is expected to denote meaningful head movement features. Then the driving audio is mapped into the center motion feature, around which different spontaneous motion features are sampled in a reasonable range. 
Second, the \emph{Expression Feature Extractor} predicts expression features from audio signal to control mouth shape where the landmark loss is designed to improve lip-sync quality. Notably, the mouth shape is highly related to the audio signal, unlike head motions. Therefore, we adopt a one-to-one mapping approach to build this module.
Finally, one source face along with the above generated features are fed into the \emph{Video Generator} during which the window of continuous frame features is utilized to guarantee the visual stability of generated videos.

Our contributions are summarized as follows: 
\begin{itemize}
	\item OSM-Net achieves \textit{one-to-many} one-shot talking head generation with natural diverse head motions. 
	\item The Audio-Motion Mapping Network establishes the clip-level motion space rather than frame-level to improve the naturalness and manages to sample various motion features corresponding to single driving audio. 
	\item Accurate mouth shape and stable talking heads are generated through loss restriction and sequence feature input. 
	\item The state-of-the-art performance is achieved on LRW, VoxCeleb2 and HDTF dataset in terms of visual quality, lip-sync quality and generation diversity.
\end{itemize}

\section{Related Work}
\subsection{Speech-driven Talking Head Generation}
Speech-driven talking head generation methods synthesize talking head videos of source subjects under the guidance of the driving audio signal. It can mainly be divided into two categories: speaker-specific methods and speaker-independent methods. The speaker-specific methods \cite{ suwajanakorn2017synthesizing, fried2019text, zhang2021facial, lahiri2021lipsync3d, guo2021ad} can generate videos of specific speakers used in the training process.
Suwajanakorn  \etal~\cite{suwajanakorn2017synthesizing} infer the mouth area landmarks of Obama given driving audio and merge generated mouth area with fine texture into the original talking head video.
Guo \etal~\cite{guo2021ad} utilize NeRF-based~\cite{mildenhall2021nerf} network to model the head and torso separately. However,  the above methods require minutes to hours high-quality videos of each subject and merely generate a fixed range of identities, thus severely limiting the application and generalization.

Therefore, some recent works~\cite{chen2018lip, zhou2019talking, chen2019hierarchical, prajwal2020lip, chen2020talking, zhou2021pose, liang2022expressive,  zhang2021flow, wang2021audio2head, wang2022one} try to relieve the identity restrictions and utilize one source face to perform speaker-independent talking head generation. 
 Chen~\etal~\cite{chen2018lip} first explore to generate lip-sync mouth area movement. They design a multi-stream GAN-based network and propose a correlation loss between two modalities.
Later, Chen~\etal~\cite{chen2019hierarchical} design a dynamical pixel-wise loss and regression-based discriminator to predict facial landmarks from audio and generate inner talking faces.  Prajwal~\etal~\cite{prajwal2020lip} utilize a carefully-designed pre-trained lip-sync discriminator to improve the lip synchronization quality. 
{\color{black}
Yu~\etal~\cite{yu2020multimodal} utilize time-delayed LSTM to predict landmarks of mouth area, fuse them into facial landmarks of source video and generate final talking heads.
Song~\etal~\cite{song2022audio} adopt an audio-visual Style Translation Network and a Retrieval-based Video Renderer to perform video dubbing on limited video data. Nevertheless, these works simply edit the mouth region and keep the other areas of faces static. Their performance is unnatural because of the fixed facial contour, blending traces around the mouth, and no head motion changes.
}

To generate more natural talking head videos, current one-shot methods turn to produce natural head motions~\cite{ zhou2021pose, zhang2021flow, wang2021audio2head, wang2022one, wang2022progressive}. 
{\color{black}
Some methods~\cite{ zhou2021pose, wang2022progressive,liu2023opt} utilizes the auxiliary reference head movement video to guide the head motion of the generated talking heads. They directly extract head motion feature from the provided head audio. However, these methods deviate from the one-shot setting and bring a burden to the practical applications like fast short video creations since users have to find and upload another high-quality video. Later methods explore to render head motion from the audio signal to alleviate the problem. 
}
Wang \etal~\cite{wang2021audio2head} design a motion-aware recurrent neural network to predict head motions and utilize flow-based model~\cite{siarohin2019first} to generate new images.
Zhang \etal~\cite{zhang2021flow} predict mouth, eyebrow and head pose parameter sequences based on audio and propose a flow-guided network to achieve high-resolution synthesis. Wang \etal~\cite{wang2022one} design an audio-visual correlation transformer that takes phonemes and facial keypoints as input to predict dense motion fields.  All of the above methods simply perform the one-to-one mapping between audio signals and head motion sequences. Their paradigm introduces ambiguity since people behave differently in head motions when speaking the same content, resulting in an unnatural performance.

\subsection{Video-driven Talking Head Generation}
Video-driven works animate talking heads guided by other videos rather than audio signals, which is also called face reenactment. According to different types of driving face representation, face reenactment can be roughly divided into four categories, facial landmark-based, 3D model-based, motion field based, and feature-based methods. Facial landmark-based models~\cite{ha2020marionette, zhang2020freenet, liu2021li, ren2023hr} utilize pre-trained landmark detector~\cite{bulat2017far, zhang2017detecting} to predict facial landmarks of driving faces and employ image translation network~\cite{pix2pix2017, wang2018high} to produce new talking heads. 3D model-based methods~\cite{thies2016face2face, shen2018faceid, nagano2018pagan, yao2020mesh, ren2021pirenderer,peng2021unified} utilize 3D morphable model~\cite{blanz1999morphable, deng2019accurate} to separate identity, expression, pose parameters and render new faces into the image plane. Instead of using explicit facial structure representation, the motion filed based methods~\cite{siarohin2019animating, siarohin2019first, hong2022depth, zhao2022thin} try to predict motion flow to represent the variation in each part of the source face image and design an occlusion-aware generator to reenact the source faces. Other methods~\cite{zakharov2019few, burkov2020neural, wang2020imaginator, zeng2020realistic} turn to disentangle compressed identity and expression features from source face and driving videos and employ the StyleGAN~\cite{karras2019style, karras2020analyzing} architecture to perform face reenactment.  Recent methods~\cite{wang2021one, chen2023compact} also explore to extract global facial feature representation for talking head video frames and perform the video compression.
Among the above methods, 3D model-based methods can achieve natural explicit pose control and better expression-identity disentanglement.

{\color{black}

\subsection{Memory Network}
The Memory Network~\cite{weston2014memory} first constructs a long-term memory network that can be read and written by users with inference capability. Later, Key-Value Memory Network~\cite{miller2016key} is proposed where key memory is utilized to locate query information and the value memory is addressed through the query. The structure is efficient to store related knowledge and information. Yi~\etal~\cite{yi2020audio} adopt memory networks into talking head generation, aiming to fine tune rendered frames into realistic frames. Jiang~\etal~\cite{jiang2022text} use memory networks to store text and image features to perform the text-to-image task. In OSM-Net, inspired by previous works, we construct the motion space to build the mapping between audio signals and head pose features.

}

\section{Method}
\begin{figure*}[t]
	\centering
	\includegraphics[width=\textwidth]{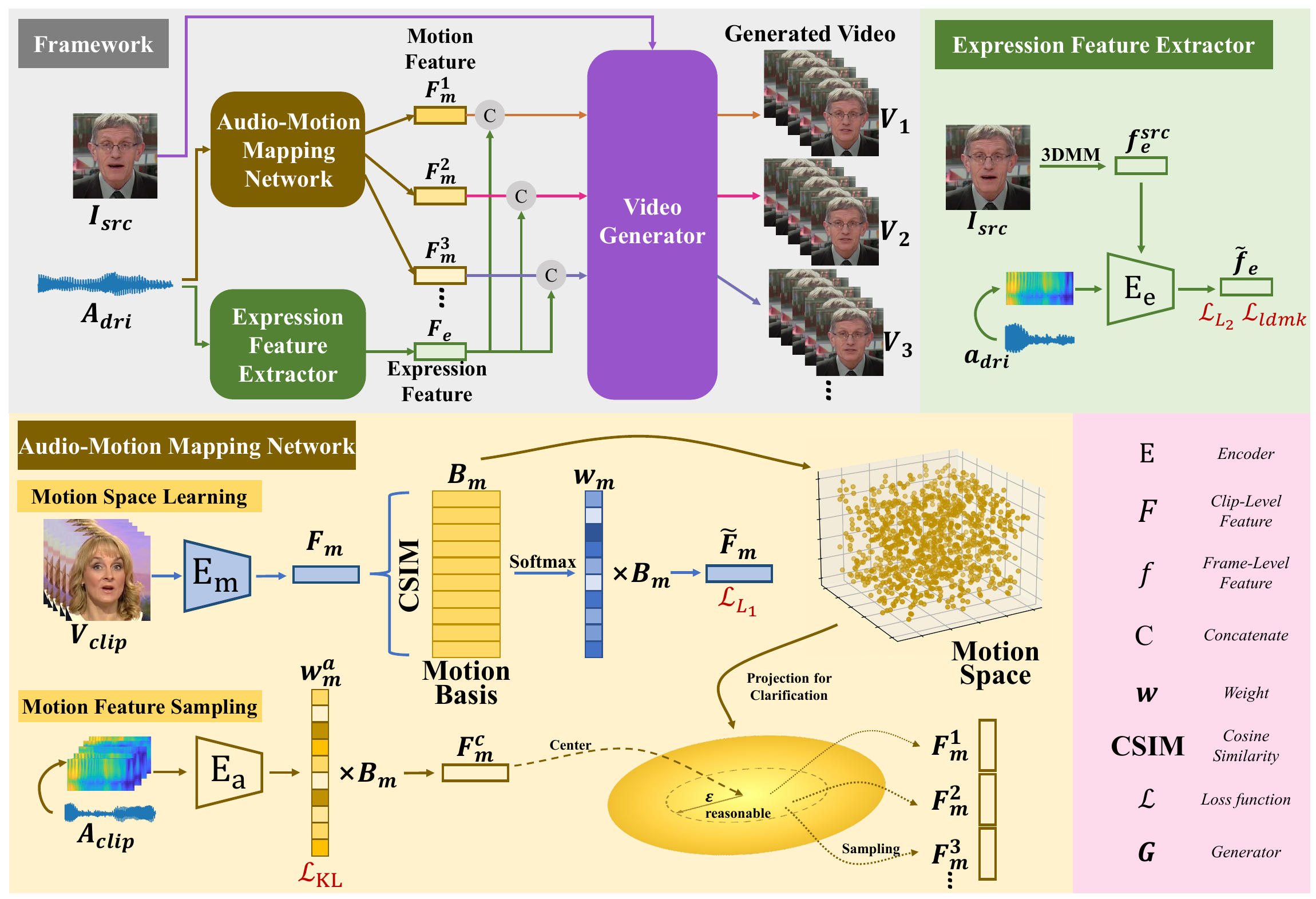} 
	\caption{{Framework and model details of OSM-Net.} 
		OSM-Net extracts expression features and diverse motion features from audio to achieve one-to-many talking head generation with diverse spontaneous head motions.
		The \emph{Audio-Motion Mapping Network} learns multiple spontaneous motions from audio. The basis of motion space $B_{m}$ is learned through the reconstruction of clip-level motion features $F_{m}$ extracted from video clip $V_{clip}$. Then the center motion feature $F^{c}_{m}$  of the corresponding audio clip $A_{clip}$ is mapped into the motion space. Multiple spontaneous motion features $F^{n}_{m}$ are sampled around $F^{c}_{m}$ within a fixed range $\epsilon$. The \emph{Expression Feature Extractor} extracts frame-level facial expression features $\tilde{f}_{e}$ from audio frame $a_{dri}$ and $I_{src}$. 
		Finally, combined motion and expression features are assembled together and sent to the \emph{Video Generator} together with $I_{src}$ to generate the final talking head.  }
	\label{fig:framework_detail}
\end{figure*}

OSM-Net aims to achieve one-to-many one-shot talking head generation with diverse spontaneous head motions. The overall framework is depicted in Fig. \ref{fig:framework_detail}. Motion features and expression features are extracted from driving audio and fed into the Video Generator along with one source face. In this section, we introduce each module in detail.

\subsection{Audio-Motion Mapping Network} 

In this module, we aim to build the one-to-many mapping between audio and motion features. Previous methods either rely on the reference video or directly build the one-to-one mapping between two modalities. However, in the Audio-Motion Mapping Network, we explore to establish the diverse one-to-many mapping.
Given the fact that people may perform different natural head motions even when speaking the same content, we believe there exists a head motion space that describes all kinds of reasonable head motions. Any driving audio clip can be mapped into part of the motion space to perform the one-to-many mapping form. Specifically, the Motion Space Learning Module learns motion basis from massive real talking head video clips and constructs the natural motion space. Each motion basis is expected to denote meaningful head movements. The Motion Feature Sampling Module then builds the mapping between audio clips to diverse clip-level motion features.

\subsubsection{Motion Space Learning}
Unlike previous methods which merely focus head motion in a single frame, OSM-Net explores clip-level motion features to model the continuous motion changes.
To achieve this,  OSM-Net learns a universal clip-level motion space from face video clips. 
The clip-level motion space is represented by a set of motion basis denoted as $B_{m} = \{ b_{m}^{i}\}_{i=1}^{S}$ where $b_{m}^{i} \in \mathbb{R}^{C}$ is clip-level motion basis learned from video clips. Different weighted combinations of motion basis yield different motion features.
Specifically, let $V_{clip}$ denote a face video clip containing $t$ consecutive frames, and $F_m$ represent its corresponding motion features extracted by motion encoder $E_m$.
When $F_m$ is given as a query, the distance between $F_m$ and each of the basis is computed using cosine similarity:

\begin{equation}
	d_{m}^{i}=\frac{b_{m}^{i} \cdot F_{m}}{\left\|b_{m}^{i}\right\|_{2} \cdot\left\|F_{m}\right\|_{2}}.
\end{equation}

Then we take Softmax operation on $d_{m}^{i}$ to calculate the attention value $\alpha_{m}^{i}$ on the i-th basis with respect to $F_m$, where $\kappa$ is a scaling term:

\begin{equation}
	\alpha_{m}^{i}=\frac{\exp \left(\kappa \cdot d_{m}^{i}\right)}{\sum_{j=1}^{S} \exp \left(\kappa \cdot d_{m}^{j}\right)}.
\end{equation}

By computing the attention value for all $b_{m}$, we get the weight $w_{m} = \{ \alpha_{m}^{1}, \alpha_{m}^{2}, ... \alpha_{m}^{S} \} \in \mathbb{R}^{S}$. It indicates the relevance between clip-level motion feature $F_{m}$ and each of the basis $b_{m}$. Then we reconstruct $F_{m}$ by taking dot product between weight and basis: $\tilde{F}_{m} = w_{m} \cdot B_{m}$. To store meaningful clip-level motion features into basis, the $L1$ loss is utilized between $F_{m}$ and $\tilde{F}_{m}$:

\begin{equation}
	\mathcal{L}_{basis}=\left\|F_{m}-\tilde{F}_{m}\right\|_{1}.
\end{equation}

In this way, each motion basis contains meaningful clip-level motion features learned from various video clips.  To consolidate the effect of feature $F_{m}$ and each motion basis about controlling the head motion, during training $F_{m}$ is fed into the Video Generator to guide the generation process. Please see Section \ref{sec_generator} for details.

\subsubsection{Motion Feature Sampling}
After building the motion space based on the motion basis learned above, we aim to seek the principle of the one-to-many mapping between driving audio and diverse head motion sequences. During training, given audio clip $A_{clip}$ that corresponds to $V_{clip}$, the audio encoder $E_a$ is adopted to predict the weight $w_{m}^{a}$ that indicates the relevance concerning each motion basis. The KL divergence loss is utilized to align weights extracted from image and audio modality:

\begin{equation}
	\mathcal{L}_{\text {KL }}=D_{K L}\left(w_{m} \| w_{m}^{a}\right).
\end{equation}

In this way, we successfully build the mapping between visual and audio modalities.  To dig out the reasonable head motion sequences corresponding to the audio signal, we aim to find a suited distribution $\mathcal{F}$ in the motion space. Concretely, during testing, for each driving audio clip $A_{dri}^{clip}$, the corresponding weight can be obtained from $E_a$. Hence, we get the clip-level center motion feature $F_{m}^{c}$ by dot product through each motion basis.  $F_{m}^{c}$ can be seen as the center of the desired reasonable distribution $\mathcal{F}$. Then, the desired reasonable motion feature distribution $\mathcal{F}$ corresponding to current $A_{dri}^{clip}$ is defined as:
\begin{equation}
	\mathcal{F} = \left \{ F | \| F-F_{m}^{c} \|_{2} \leq \epsilon\right\},
\end{equation}
where $\epsilon$ is the distance to locate the reasonable scope of motion space. Different motion features can then be sampled from $\mathcal{F}$ to represent spontaneous clip-level motions corresponding to current $A_{dri}^{clip}$. 

After getting consecutive clip-level motion features corresponding to a long driving audio, we form them as holistic video-level motion features. It is worth mentioning that each feature in the motion space denotes the \textit{difference} corresponding to the first frame, not the ground truth value. The main reason is that the initial head motion states in different video clips are different. When combining several clip-level motion features, the last head motion state of the previous clip-level motion features serves as the initial state of the next ones. In this way, the holistic video-level motion features are formed and further fed into the Video Generator to guide the head motion control.

\subsection{Expression Feature Extractor}
After obtaining motion features, we aim to explore expression features that describe facial movement and mouth shape. To focus more on the inner mouth shape accuracy within talking head frame, the \emph{frame-level} expression feature representation is adopted in OSM-Net. 

{\color{black}
In the one-shot talking head generation task, the mouth shape and head motion features should both be inferred from the driving audio. The mouth shape is highly related to the audio signal since the content of speech is directly displayed by the mouth area, leading to the one-to-one mapping. However, the head motion feature has only weak correlations with the phonetic signal. The same person can behave differently in head motions when speaking the same content. Hence, the one-to-many mapping is adopted in the Audio-Motion Mapping Network. In this section, we propose to utilize a separated one-to-one mapping module called the Expression Feature Extractor to predict expression features that describe facial movement and mouth shape.
}

Considering the good reconstruction performance of 3D Morphable Model\cite{blanz1999morphable, deng2019accurate}, we use a subset of 3DMM parameters to represent the expression feature.
With a 3DMM, the 3D shape $\mathbf{S}$ of a face can be parameterized using following formula:
\begin{equation}
	 \mathbf{S}=\overline{\mathbf{S}}+\boldsymbol{\alpha} \mathbf{B}_{i d}+\boldsymbol{\beta} \mathbf{B}_{e x p},
\end{equation}
where $\overline{\mathbf{S}}$ is the average face shape, $\mathbf{B}_{i d}$ and $\mathbf{B}_{e x p}$ are basis of identity and expression via PCA based on scans of human faces \cite{paysan20093d}. The coefficients $\alpha$ and $\beta$ describe the facial shape and expression respectively. In the Expression Feature Extractor,the coefficient $\beta$ is chosen as the demanding expression feature  $f_e$.

Specifically, as shown in green part of Fig. \ref{fig:framework_detail}, given frame-level driving audio $a_{dri}$ and source face $I_{src}$, the Expression Feature Extractor aims to predict expression feature $f_e$ corresponding to $a_{dri}$. The following losses are adopted: 

\begin{equation}
	\mathcal{L}_{exp} = \mathcal{L}_{L2} + \lambda_{ldmk}\mathcal{L}_{ldmk}.
\end{equation}

The $L2$ loss $\mathcal{L}_{L2}$  guarantee the accuracy on feature level:
\begin{equation}
	 \mathcal{L}_{L2} = \left\|   \tilde{f}_e  -  f_{e} \right\|_{2}.
\end{equation}
In order to further improve the accuracy of mouth shape and facial structure behind the expression coefficient $\beta$ in 3DMM, we design the facial landmark loss $\mathcal{L}_{ldmk}$ using 3DMM:

\begin{equation}
	\mathcal{L}_{ldmk}=\frac{1}{N} \sum_{n=1}^{N} \omega_{n}\left\|\tilde{\mathbf{p}}_{n}-\mathbf{p}_{n}\right\|^{2}
\end{equation}
where $\{\mathbf{p}_{n}\}$ is facial landmarks of ground truth driving face image, $\{\tilde{\mathbf{p}}_{n}\}$ is the 3D landmark vertices projection of reconstructed shape using $f_e$ onto the image plane. $N$ denotes the number of landmarks. The weight $\omega_{n}$ id set to 20 for inner mouth and nose points and others to 1. During testing, the video-level expression features with respect to any driving audio can be obtained frame by frame.

\subsection{Video Generator} \label{sec_generator}
After obtaining video-level motion and expression features, the Video Generator aims to generate natural talking head videos. For each frame, we concatenate the corresponding motion and expression feature to form its holistic driving feature. For any given driving audio that corresponds to $T$ talking head video frames, $T$ driving features are obtained.

\begin{figure}[t]
	\centering
	\includegraphics[width=\columnwidth]{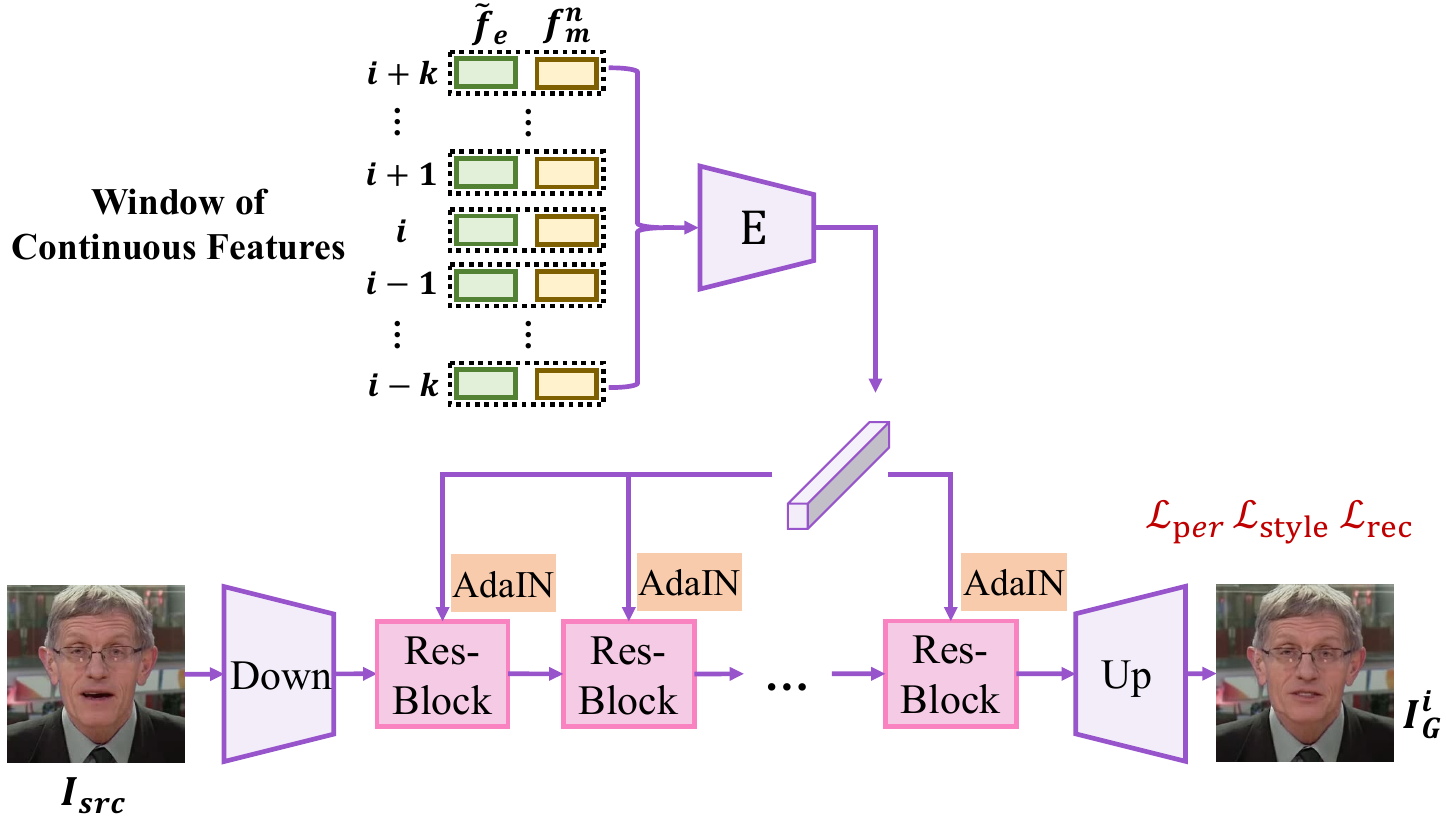} 
	\caption{{Details of Video Generator.} The window of combined frame-level expression and motion features $\{\tilde{f}_{e} , f^{n}_{m}\}_{i-k}^{i+k}$ are assembled together and sent to the encoder to get high level features. Then the conditional auto-encoder structure is utilized to translate $I_{src}$ into final $i_{th}$ talking head image $I_{G}^{i}$.  }
	\label{fig:gen}
\end{figure}

\begin{figure*}[t]
	\centering
	\includegraphics[width=\textwidth]{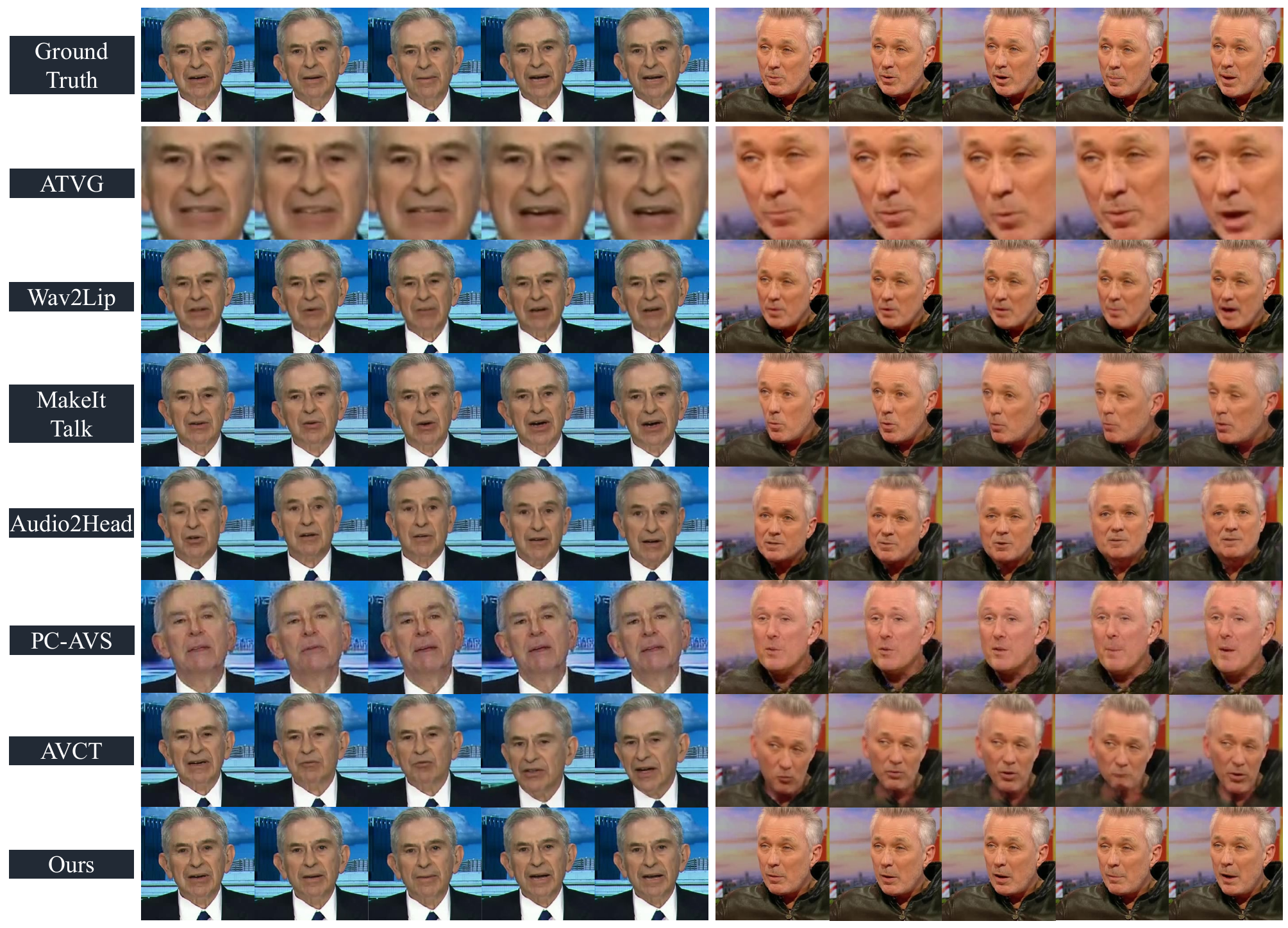} 
	\caption{{Qualitative comparisons with other state-of-the-art methods.} Note the mouth shape and facial contour of generated faces.(Zoom in for details.)  }
	\label{fig:quality}
\end{figure*}

Furthermore, to improve the stability of generated face videos and alleviate the impact led by the estimation errors from 3D face reconstruction, we utilized features corresponding to a window with continuous frames to generate the face in the middle of the window. In this way, the network can be expected to avoid errors by extracting relations between adjacent frames. 

In detail, as shown in Fig.~\ref{fig:gen}, the Video Generator contains two parts and requires an end-to-end training. First, the window of motion and expression features will be fed into an encoder based on several 1D convolution blocks to extract high-level 256-dim features, during which the residual link with CenterCrop operation is added around each convolution part. The CenterCrop operation will gradually increase the focus of the middle feature while taking the context feature information into account. Then we design the generation part based on the conditional auto-encoder structure~\cite{isola2017image}. The source images are first fed into three convolution layers with average pooling and then go through five common residual layers with feature map size unchanged. Finally, the feature map is up-sampled with three residual blocks with the transpose convolution layer to output final talking face images. Concatenation-based skip connections with convolution operation are employed to skip the high-resolution features. The 256-dim features extracted above are injected after each Convolution and Transpose Convolution parts using AdaIN~\cite{huang2017arbitrary} operation. The layer normalization is used as the activation normalization of each convolution layer, while the Leaky-ReLU is adopted as the non-linear function in the generator. Overall, the whole generation process of $i_{th}$ frame is formulated as:

\begin{equation}
	\tilde{I}_G^{i} = G \left( H^{i}_{combine} \left(F_{m}, F_{e}\right) , I_{src}\right).
\end{equation}

\begin{equation}
	\tilde{I}_g^{i} = G \left( H^{i}_{combine} \left(F_{m}^{c}, F_{e} \right) , I_{src}\right),
\end{equation}
where $H_{combine}^i$ denotes a length of $2k+1$ window of concatenated motion and expression features with the $i_{th}$ features in the middle, as shown in the purple part of Fig. \ref{fig:framework_detail}. As for the network details, consecutive driving features will first be mapped into driving vectors using 1D Convolution and residual link with CenterCrop. 

During training the reconstruction loss $\mathcal{L}_{ rec }$, the perceptual loss $\mathcal{L}_{per}$ and style loss $\mathcal{L}_{style}$ are adopted. The $\mathcal{L}_{ rec }$ is the $L1$ loss imposed on the above generated frames and ground truth frames $I$:

\begin{equation}
	\mathcal{L}_{rec} = \left(\left\| \tilde{I}_{g} - \left(I\right)\right\|_{1} + \left\| \tilde{I}_{g} - \left(I\right)\right\|_{1} \right)
\end{equation}

 $\mathcal{L}_{per}$ calculates the distance between activation maps of the pre-trained VGG-19 network :

\begin{equation}
	\resizebox{\columnwidth}{!}{$
		\mathcal{L}_{per}=\sum_{i}\left(\left\|\phi_{i}(\tilde{I}_{g})-\phi_{i}\left(I\right)\right\|_{1} + \left\|\phi_{i}(\tilde{I}_{G})-\phi_{i}\left(I\right)\right\|_{1} \right),
		$}
\end{equation}
where $\phi_{i}$ is the activation map of the i-th layer of the VGG-19 network \cite{simonyan2014very}. $\mathcal{L}_{style}$ calculates the difference between Gram matrix constructed from VGG-19 activation maps $\phi_{j}$:

\begin{equation}
	\resizebox{\columnwidth}{!}{$
		\mathcal{L}_{style}=\sum_{j} \left( \left\|G_{j}^{\phi}(\tilde{I}_{g})-G_{j}^{\phi}(I)\right\|_{1} + \left\|G_{j}^{\phi}(\tilde{I}_{G})-G_{j}^{\phi}(I)\right\|_{1}
		\right).
		$}
\end{equation}

The final loss of Video Generator is a summation of the above losses:

\begin{equation}
	\mathit{L} = \lambda_{rec}\mathit{L}_{rec} + \lambda_{per}\mathit{L}_{per} + \lambda_{style}\mathit{L}_{style}.
\end{equation}

During testing, various video-level motion features can be sampled from reasonable motion space. Different combinations of motion and expression features lead to various talking head videos with natural head motions, performing one-to-many mapping.

\begin{table}[t]
	\caption{{Quantitative comparisons with state-of-the-art methods on LRW dataset. The {bold} and {underlined} notations represents the Top-2 results.  }}
	\centering
	\setlength\tabcolsep{3pt}
	\resizebox{\columnwidth}{!}{
		\begin{tabular}{c c c c c c}
			\toprule
			Method & SSIM $\uparrow$ & CPBD $\uparrow$ & LMD $\downarrow$  & LSE-C $\uparrow$  & Diversity $\uparrow$ \\
			\midrule
			ATVG    &0.781   &0.102  &5.25     &4.165&N./A. \\
			Wav2Lip      & \underline{0.812}  & 0.152  &5.73     &\textbf{7.237} & N./A.\\
			MakeItTalk    & 0.796  & 0.161 &7.13 &3.141& \underline{0.082 }\\
			Audio2Head    &0.743   &0.168  &7.34  &2.135 &0.064 \\
			PC-AVS         & 0.778  &\textbf{0.185}  &3.93    &6.420&N./A. \\
			AVCT         &0.805   & 0.181  & \underline{3.56}  &6.567&0.059 \\
			Ground Truth     &1.000   &{0.173}  &0.00     &6.876& 0.103\\
			Ours             &\textbf{0.831}   & \underline{0.182}  &\textbf{3.13}    & \underline{6.583} &\textbf{0.115}\\
			\bottomrule
		\end{tabular}
	}
	
	\label{tab:LRW}
\end{table}

\begin{table}[t]
	\caption{{Quantitative comparisons with state-of-the-art methods on HDTF dataset.   }}
	\centering
	\setlength\tabcolsep{3pt}
	\resizebox{\columnwidth}{!}{
		\begin{tabular}{c c c c c c}
			\toprule
			Method & SSIM $\uparrow$ & CPBD $\uparrow$ & LMD$\downarrow$ & LSE-C $\uparrow$ & Diversity $\uparrow$   \\
			\midrule
			ATVG    &0.735   &  0.083&6.78     &5.584 &N./A. \\
			Wav2Lip      &\underline{0.776}   & 0.175 &6.53     &\textbf{8.797}  &N./A.\\
			MakeItTalk    &0.723   &0.135 &8.91  &4.356 & \underline{0.252}\\
			Audio2Head    &0.689   &0.112  &9.03 &4.938 &0.192 \\
			PC-AVS         &0.747   & 0.178  &\underline{5.68}    &6.728&N./A.\\
			AVCT         &  0.769 & \underline{0.180}  & 6.12  & 7.013&0.239 \\
			Ground Truth     &1.000   &0.192&0.00   &8.247& 0.263\\
			Ours              &\textbf{0.790}   & \textbf{0.183}&\textbf{5.45}   &\underline{7.424} & \textbf{0.317}  \\
			\bottomrule
		\end{tabular}
	}
	
	\label{tab:HDTF}
\end{table}

\begin{table}[t]
	\caption{{Quantitative comparisons with state-of-the-art methods on VoxCeleb2 dataset.   }}
	\centering
	\setlength\tabcolsep{3pt}
	\resizebox{\columnwidth}{!}{
		\begin{tabular}{c c c c c c}
			\toprule
			Method & SSIM $\uparrow$ & CPBD $\uparrow$ & LMD $\downarrow$  & LSE-C $\uparrow$ & Diversity $\uparrow$   \\
			\midrule
			ATVG    &0.826   & 0.061  &\underline{6.49}    &4.319& N./A.\\
			Wav2Lip      &0.846   &  0.078 &12.26     &\textbf{7.640}& N./A.\\
			MakeItTalk    &0.817   &   0.068&31.44  &3.756 & \underline{0.472}\\
			Audio2Head    & 0.782 &0.072   & 9.72 &3.437 & 0.206 \\
			PC-AVS         &\textbf{0.886}   &  0.083&6.88  & 5.235&N./A.\\
			AVCT         & 0.832   & 0.080  & 8.32    &3.441& 0.254\\
			Ground Truth     &1.000   &0.090 &0.00   &6.209& 0.485\\
			Ours              &\underline{0.853}  &\textbf{0.085 }&\textbf{6.32}     &\underline{6.134}   &\textbf{0.513}\\
			\bottomrule
		\end{tabular}
	}
	
	\label{tab:VoxCeleb2}
\end{table}

\section{Experiment}
\subsection{Experimental Settings}

\subsubsection{Dataset}
We quantitatively and qualitatively evaluate our method on LRW \cite{chung2016lip},  HDTF \cite{zhang2021flow} and VoxCeleb2~\cite{chung2018voxceleb2} datasets. The LRW dataset contains over 1000 utterances of 500 different words. The VoxCeleb2 contains a total of  6112 celebrities covering over a million utterances. It contains large pose movements and different light conditions. HDTF contains high resolution and in-the-wild talking head videos.

\subsubsection{Comparison Methods}
Several state-of-the-art one-shot talking head generation methods are utilized as comparing methods.
\begin{itemize}
	\item  \textbf{ATVG}~\cite{chen2019hierarchical} uses the 2D facial landmarks and an attention mechanism. 
	\item \textbf{Wav2Lip}~\cite{prajwal2020lip}  is a reconstruction-based method that focuses on editing the mouth shape to improve the lip-sync quality.
	\item \textbf{MakeItTalk}~\cite{zhou2020makelttalk} predicts content-based and identity-based 3D facial landmarks.
	\item \textbf{Audio2Head}~\cite{wang2021audio2head} infers unique head pose sequences from audio and utilizes a flow-based generator.
	\item \textbf{PC-AVS}~\cite{zhou2021pose} extracts modularized audio-visual representations of identity, pose and speech content.
	\item \textbf{AVCT}~\cite{wang2022one} designs an audio-visual correlation transformer that takes phonemes and motion fields as inputs and predicts new dense motion fields.
	\item \textbf{Ground Truth} results are also displayed as comparison.
\end{itemize}

\subsubsection{Implementation Details}
The face video frames are cropped to $256\times256$ size at 25 FPS  and audio to mel-spectrogram of size $16 \times 16$ per frame. The length of video clip $T$ is set to 5 and the number of motion basis $S$ is designed to 48 with clip-level motion feature dimension $C$ 512. The window length k and the distance $\epsilon$ are set to 8 and 1 empirically.  $E_m$ contains a pre-trained pose estimation network~\cite{deng2019accurate} to predict the transformation matrix and a pose feature encoder to extract $C$ dimension feature. In the Video Generator, a decoder is first adopted to map motion features into transformation matrix $[R, T] \in \mathbb{R}^{6}$ where $R$ denotes the rotation matrix and  $T$ denotes the translation value.

As for training, our method is trained in stages. The Hyper-parameters are empirically set: $\lambda_{rec}$ to 4, $\lambda_{pre}$ to 2.5, $\lambda_{style}$ to $10^3$ and $\lambda_{ldmk}$ is set to 0.02.  The ADAM optimizer is adopted with an initial leaning rage as $10^{-4}$. The learning rate is decreased to $2 \times 10^{-5}$ after $300k$ iterations. We train on 4 Tesla 32G V100 GPUs and Intel Xeon Gold CPU.

\subsection{Quantitative Evaluation}

\subsubsection{Evaluation Metrics}
We evaluate the performance on three aspects, image quality, lip-sync quality and generation diversity. The SSIM \cite{wang2004image} and Cumulative Probability of Blur Detection (CPBD)~\cite{narvekar2009no} scores are utilized to judge the talking head image quality. For the lip-sync quality, the Landmark Distance (LMD)~\cite{chen2018lip} and Lip-Sync Error-Confidence (LSE-C) are applied. LMD means the average Euclidean distance between facial landmarks of generated faces and corresponding ground truth faces.  LSE-C is the confidence score of the correspondence between audio and video features extracted from pre-trained SyncNet \cite{chung2016out}. 
{\color{black}
To evaluate the generation diversity, the standard deviation of the head motion feature sequence is calculated using pose estimation network Hopenet~\cite{ruiz2018fine}.
}

\begin{table}[t]
	\caption{{User study results by mean opinion scores.}}
	\centering
	\resizebox{\columnwidth}{!}{
		\begin{tabular}{c c c c c}
			\toprule
			Method & Visual  & Lip-Sync  & Identity & Diversity \\
			\midrule
			Ground Truth  &  4.72  & 4.38   & 4.90    & N./A. \\
			ATVG       &  2.30  &3.26  & 2.59  & 1.72  \\
			Wav2Lip       &  3.43  &3.87  & 3.15  & 1.49  \\
			MakeItTalk    & 3.79 &2.69  & 3.41  & 3.56  \\
			Audio2Head    &3.67  & 2.78 &  3.72&  3.10 \\
			PC-AVS        &  3.75&3.68  & 2.96  & 1.86  \\
			AVCT       &  3.92& 3.91 & 3.68  &3.23   \\
			
			Ours          &  \textbf{4.16}& \textbf{4.31} &  \textbf{3.85} &  \textbf{4.07} \\
			\bottomrule
		\end{tabular}
	}
	
	\label{tab:user_study}
\end{table}

\subsubsection{Image Generation Paradigm}
We conduct all the experiments under self-driven one-shot setting. The first frame in the test video is selected as the source identity frame and the corresponding audio serves as driving audio. The metrics are calculated between generated video frames and corresponding ground truth video frames. Since our method could generate talking heads with diverse motions, during evaluation, we report the results generated by the guidance of the center motion feature for a fair comparison.

\subsubsection{Evaluation Results}
\Cref{tab:LRW,tab:HDTF,tab:VoxCeleb2}  show the quantitative comparison results based on LRW, HDTF and VoxCeleb2 datasets. All results are generated under one-shot setting. It shows that OSM-Net achieves leading SSIM, CPBD and LMD scores in all datasets. As for lip-sync evaluation, Wav2Lip sometimes performs the best, even better than the ground truth. Furthermore, as mentioned in \cite{zhou2021pose}, the leading LSE-C only means that Wav2Lip is comparable to the ground truth, not better. The LSE-C of OSM-Net is also close to the ground truth and our leading LMD score indicates that we preserve better facial structure and accurate mouth shape. 
{\color{black}
Finally, thanks to the Audio-Motion Mapping Network, OSM-Net achieves diverse generation on head movements and gets leading diversity scores.
}

\subsection{Qualitative Evaluation}
Given one source face, our method generates talking head videos with many diverse spontaneous motions, as shown in Fig. \ref{fig:teaser} and the supplementary material. Furthermore, we compare OSM-Net with previous SOTA methods, as displayed in Fig. \ref{fig:quality}. It shows that our method generates high-quality talking head videos with natural head motions, accurate mouth shape and facial contours that best match the ground truth. Concretely, ATVG merely focuses on the cropped facial regions. Wav2Lip generates faces with fixed head motions and blurry mouth shapes. Though MakeItTalk and Audio2Head contain head pose changes, they fail to keep lip synchronization and accurate facial contour. PC-AVS cannot preserve the identity features of source faces. As for AVCT, it performs well when source faces are nearly frontal. However, when dealing with the side face situation, it suffers a lot from the visual artifacts and identity mismatch problem, meaning the inability to preserve the identity of source faces.

 \begin{figure}[t]
 	\centering
 	\includegraphics[width=0.9\columnwidth]{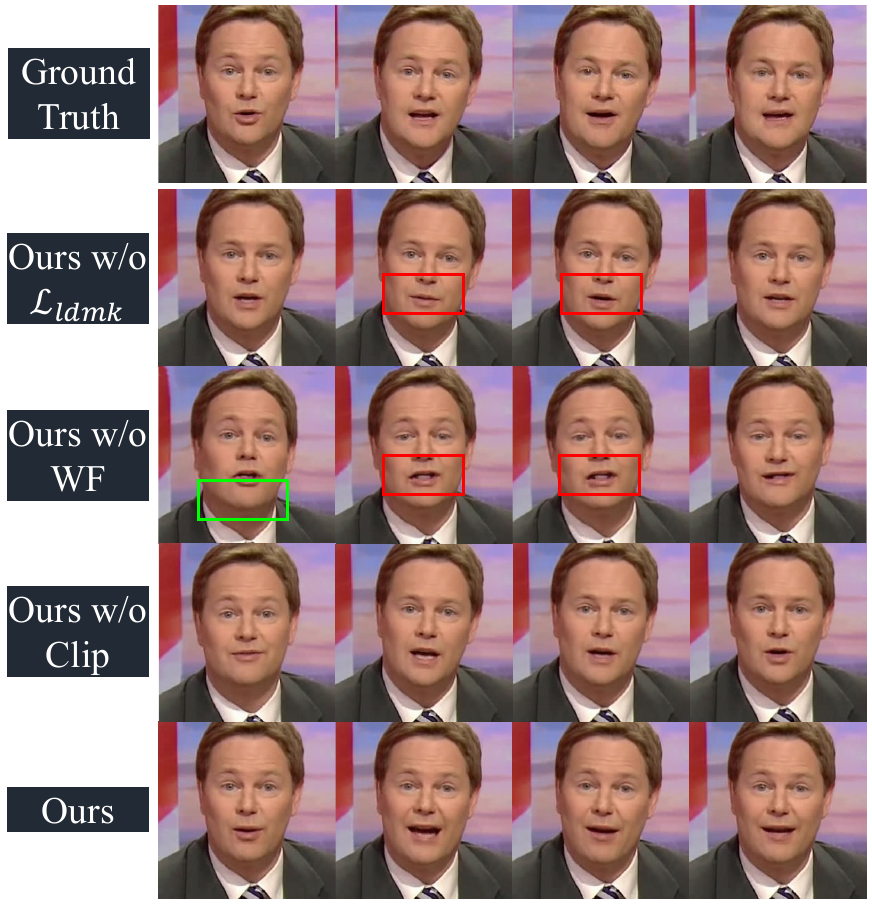} 
 	\caption{{Qualitative results of ablation study.} We show four consecutive frames. Notice the difference of mouth shape in red rectangle, facial contour in green rectangle and head motion change in Ours w/o Clip.   }
 	\label{fig:ablation}
 \end{figure}

 \begin{table}[t]
 	\centering
 	\caption{{Quantitative results of ablation study on LRW.}}
 	\resizebox{0.85\columnwidth}{!}{
 		\begin{tabular}{c c c c}
 			\toprule
 			Method & SSIM $\uparrow$ & LMD $\downarrow$ & LSE-C $\uparrow$ \\
 			\midrule
 			Ours w/o $\mathcal{L}_{l d m k}$ & 0.785 & 6.67 & 5.721 \\
 			Ours w/o WF & 0.763 & 4.02 &5.982 \\
 			Ours w/o Clip & 0.758 & 4.27 & 6.147 \\
 			Ours & \textbf{0.831} & \textbf{3.13} & \textbf{6.583} \\
 			\bottomrule
 		\end{tabular}
 	}

 	\label{tab:ablation}
 \end{table}

\subsection{User Study}
We further conduct a user study to compare the generation results. For each method, we randomly generate 25 videos for each dataset. Each video is generated repeatedly three times to verify the generation diversity. 10 participants are asked to grade the 150 generated videos based on visual quality, lip-sync quality, generation diversity and identity preservation . The popular Mean Opinion Scores (MOS) are adopted, where a score between 1 to 5 is given. As shown in Table \ref{tab:user_study}, OSM-Net achieves the best results in all aspects, especially generation diversity. The results indicate the  superiority of the one-to-many mapping in OSM-Net.

\subsection{Further Analysis} 

\subsubsection{Ablation Study}
To evaluate the effectiveness of designed components in OSM-Net, we conduct the ablation study on the \emph{landmark loss $\mathcal{L}_{ldmk}$}, the \emph{window of continuous features (WF)} and whether to consider the \emph{clip-level motion feature (Clip)}. 
Ours w/o WF utilizes a single combined feature as input to the generator and Ours w/o Clip replaces clip-level motion features with frame-level motion features. The qualitative and quantitative results are shown in Fig. \ref{fig:ablation} and Table \ref{tab:ablation}.  They both demonstrate $\mathcal{L}_{ldmk}$ is crucial to the mouth shape and $WF$ help improve the video stability. Without $\mathcal{L}_{ldmk}$, LMD and LSE-C scores suffer a huge decline and mouth shape could not match the ground truth. The red rectangle shows the inaccurate mouth shape. After introducing the WF, the image quality SSIM improves a lot. The green rectangle also shows that WF can help improve the stability of facial contour. Ours w/o Clip suffers severe motion changes between adjacent frames and causes SSIM decline. Moreover, we do not explicitly perform an ablation study on motion space learning since we could not get diverse talking head videos without the motion space construction and sampling.

\begin{figure}[t]
	\centering
	\includegraphics[width=\columnwidth]{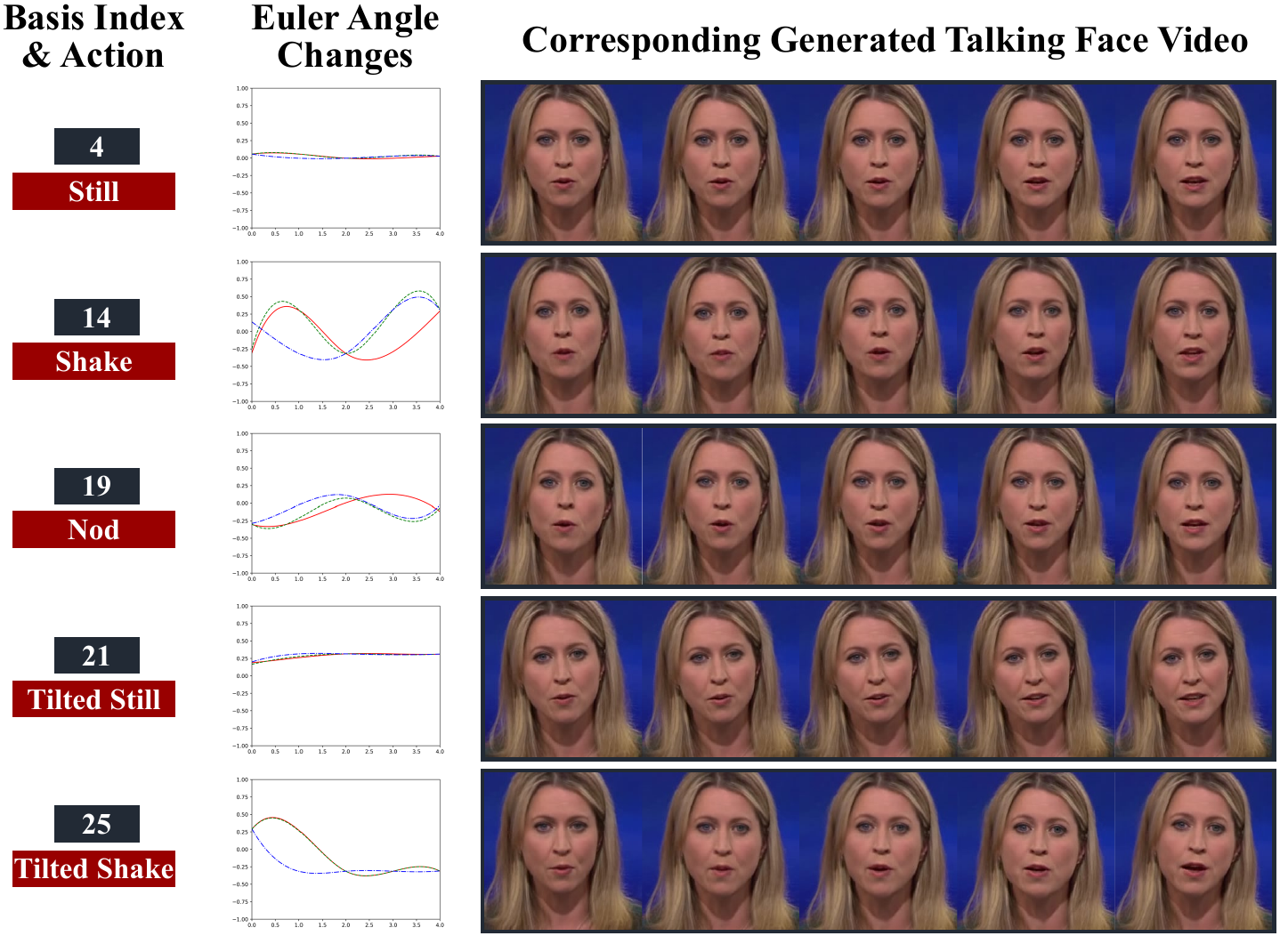} 
	\caption{{Motion Basis Analysis.} Motion basis index with the corresponding action, Euler angle changing curves of head movement and generated talking heads are displayed.   }
	\label{fig:ana_basis}
\end{figure}

\subsubsection{Motion Basis Analysis}
This section aims to further analyze the motion space and its corresponding basis. Each motion contained in the basis needs to be visualized. Therefore, a random subject is selected from the LRW dataset as the source identity and a random audio sequence is also selected to provide expression features. The motion features are provided by each basis and the results are displayed in Fig. \ref{fig:ana_basis}. It is possible to assign each basis with a specific action. For instance, basis 4 denotes the still motion, basis 19 nodding, and basis 21 tilled still motion. It demonstrates that each basis represents the motion at the action level. By taking combinations of related motion basis through weight control, multiple spontaneous motions could be generated.

\subsubsection{Face Rotation and Frontalization} 
\begin{figure}[t]
	\centering
	\includegraphics[width=\columnwidth]{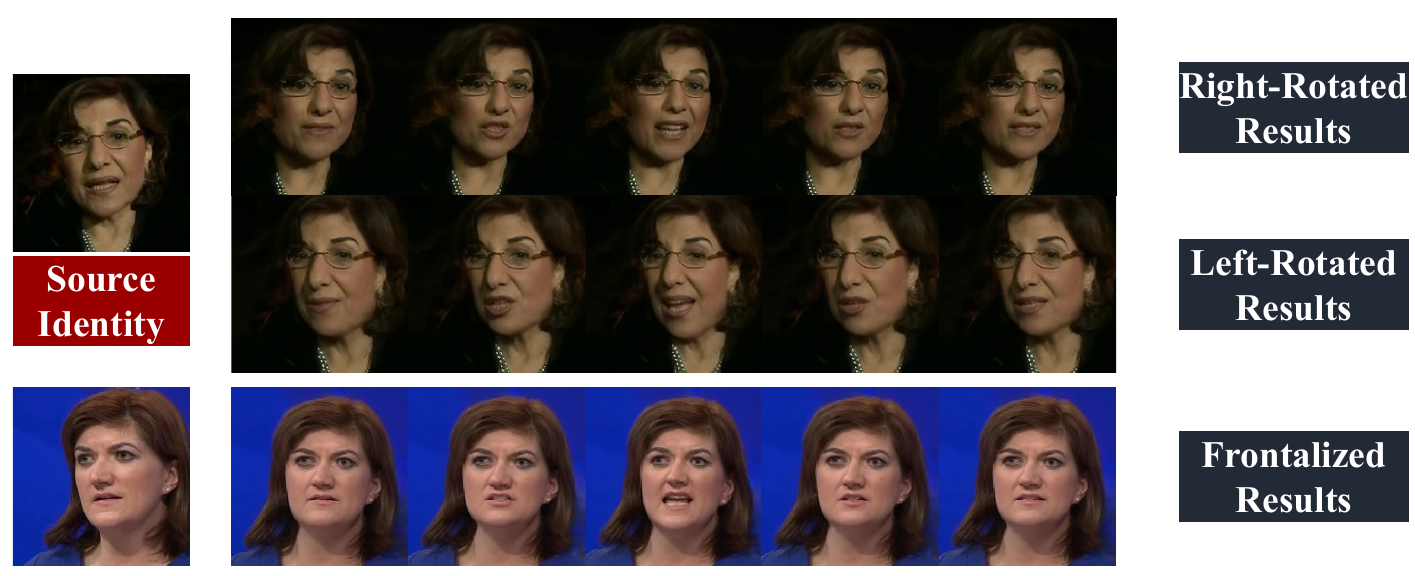} 
	\caption{{Face Rotation and Frontalization Results.} Faces of various motions can be  generated by explicitly changing the motion feature into existing motion basis.  }
	\label{fig:face_rotation}
\end{figure}

In order to verify the effectiveness of the motion basis, we design face rotation and frontalization experiments. All faces are generated under the one-shot setting, and the corresponding motion features are replaced by certain motion basis representing right-rotated, left-rotated and frontal.  Fig. \ref{fig:face_rotation} shows the generated talking heads which proves the effectiveness of motion basis.

\section{Ethical Considerations}
Our proposed method is intended for video editing industry and focuses on world-positive use cases and applications. We sincerely hope that relevant industries like film making, short video creation, metaverse, or digital avatars can benefit a lot from OSM-Net. However, we do acknowledge the potential of its misuse. Lawbreakers may spread fake videos out of malicious proposes on social media, which leads to severe negative impacts on the whole society. Therefore, we strongly support all relevant safeguarding measures against such malicious applications. We would contribute our method to the deepfake detection community and call for the enactment and enforcement of legislation to obligate all generated videos to be clearly labeled. Finally, it is necessary to obtain permission from all subjects whose faces are acted as source faces in generated talking head videos. We believe the proper usage of this technique will enhance the development of artificial intelligence research and relevant multimedia applications.

\section{Conclusion}

OSM-Net successfully performs one-to-many mapping and generates diverse sequences with natural spontaneous motions given one source face and any driving audio. It constructs a clip-level motion space through learning from a paired audio-visual corpus. The center motion feature corresponding to the audio signal is mapped into the space around which spontaneous motion features can be sampled. We verify that each motion basis denotes certain actions and a weighted combination of motion basis yields diverse spontaneous motions. Both quantitative and qualitative experiments demonstrate the superiority. 

For future work, we aim to find the relationship between the semantic information of driving speech and head motion change directions. The artifacts in the junction part of the background and the facial area also need to be eliminated.

\bibliographystyle{IEEEtran}
\bibliography{./TCSVT2023.bib}

\begin{thebibliography}{10}
\providecommand{\url}[1]{#1}
\csname url@samestyle\endcsname
\providecommand{\newblock}{\relax}
\providecommand{\bibinfo}[2]{#2}
\providecommand{\BIBentrySTDinterwordspacing}{\spaceskip=0pt\relax}
\providecommand{\BIBentryALTinterwordstretchfactor}{4}
\providecommand{\BIBentryALTinterwordspacing}{\spaceskip=\fontdimen2\font plus
\BIBentryALTinterwordstretchfactor\fontdimen3\font minus
  \fontdimen4\font\relax}
\providecommand{\BIBforeignlanguage}[2]{{%
\expandafter\ifx\csname l@#1\endcsname\relax
\typeout{** WARNING: IEEEtran.bst: No hyphenation pattern has been}%
\typeout{** loaded for the language `#1'. Using the pattern for}%
\typeout{** the default language instead.}%
\else
\language=\csname l@#1\endcsname
\fi
#2}}
\providecommand{\BIBdecl}{\relax}
\BIBdecl

\bibitem{zhang2021flow}
Z.~Zhang, L.~Li, Y.~Ding, and C.~Fan, ``Flow-guided one-shot talking face
  generation with a high-resolution audio-visual dataset,'' in
  \emph{Proceedings of the IEEE/CVF Conference on Computer Vision and Pattern
  Recognition}, 2021, pp. 3661--3670.

\bibitem{prajwal2020lip}
K.~Prajwal, R.~Mukhopadhyay, V.~P. Namboodiri, and C.~Jawahar, ``A lip sync
  expert is all you need for speech to lip generation in the wild,'' in
  \emph{Proceedings of the 28th ACM International Conference on Multimedia},
  2020, pp. 484--492.

\bibitem{kim2019neural}
H.~Kim, M.~Elgharib, M.~Zollh{\"o}fer, H.-P. Seidel, T.~Beeler, C.~Richardt,
  and C.~Theobalt, ``Neural style-preserving visual dubbing,'' \emph{ACM
  Transactions on Graphics (TOG)}, vol.~38, no.~6, pp. 1--13, 2019.

\bibitem{kim2018deep}
H.~Kim, P.~Garrido, A.~Tewari, W.~Xu, J.~Thies, M.~Niessner, P.~P{\'e}rez,
  C.~Richardt, M.~Zollh{\"o}fer, and C.~Theobalt, ``Deep video portraits,''
  \emph{ACM Transactions on Graphics (TOG)}, vol.~37, no.~4, pp. 1--14, 2018.

\bibitem{zhou2018visemenet}
Y.~Zhou, Z.~Xu, C.~Landreth, E.~Kalogerakis, S.~Maji, and K.~Singh,
  ``Visemenet: Audio-driven animator-centric speech animation,'' \emph{ACM
  Transactions on Graphics (TOG)}, vol.~37, no.~4, pp. 1--10, 2018.

\bibitem{wang2021one}
T.-C. Wang, A.~Mallya, and M.-Y. Liu, ``One-shot free-view neural talking-head
  synthesis for video conferencing,'' in \emph{Proceedings of the IEEE/CVF
  conference on computer vision and pattern recognition}, 2021, pp.
  10\,039--10\,049.

\bibitem{agarwal2022compressing}
M.~Agarwal, A.~Gupta, R.~Mukhopadhyay, V.~P. Namboodiri, and C.~Jawahar,
  ``Compressing video calls using synthetic talking heads,'' \emph{arXiv
  preprint arXiv:2210.03692}, 2022.

\bibitem{zhou2021pose}
H.~Zhou, Y.~Sun, W.~Wu, C.~C. Loy, X.~Wang, and Z.~Liu, ``Pose-controllable
  talking face generation by implicitly modularized audio-visual
  representation,'' in \emph{Proceedings of the IEEE/CVF Conference on Computer
  Vision and Pattern Recognition}, 2021, pp. 4176--4186.

\bibitem{chen2020comprises}
L.~Chen, G.~Cui, Z.~Kou, H.~Zheng, and C.~Xu, ``What comprises a good
  talking-head video generation?: A survey and benchmark,'' \emph{arXiv
  preprint arXiv:2005.03201}, 2020.

\bibitem{mirsky2021creation}
Y.~Mirsky and W.~Lee, ``The creation and detection of deepfakes: A survey,''
  \emph{ACM Computing Surveys (CSUR)}, vol.~54, no.~1, pp. 1--41, 2021.

\bibitem{liujin2022}
L.~Jin, C.~Peng, W.~Xi, F.~Xiaomeng, D.~Jiao, and H.~Jizhong, ``Critical review
  of human face reenactment methods.'' \emph{Journal of Image and Graphics},
  vol.~27, no.~9, pp. 2629--2651, 2022.

\bibitem{song2018talking}
Y.~Song, J.~Zhu, D.~Li, X.~Wang, and H.~Qi, ``Talking face generation by
  conditional recurrent adversarial network,'' \emph{arXiv preprint
  arXiv:1804.04786}, 2018.

\bibitem{zhu2018arbitrary}
H.~Zhu, H.~Huang, Y.~Li, A.~Zheng, and R.~He, ``Arbitrary talking face
  generation via attentional audio-visual coherence learning,'' \emph{arXiv
  preprint arXiv:1812.06589}, 2018.

\bibitem{zhou2019talking}
H.~Zhou, Y.~Liu, Z.~Liu, P.~Luo, and X.~Wang, ``Talking face generation by
  adversarially disentangled audio-visual representation,'' in
  \emph{Proceedings of the AAAI Conference on Artificial Intelligence},
  vol.~33, no.~01, 2019, pp. 9299--9306.

\bibitem{chen2019hierarchical}
L.~Chen, R.~K. Maddox, Z.~Duan, and C.~Xu, ``Hierarchical cross-modal talking
  face generation with dynamic pixel-wise loss,'' in \emph{Proceedings of the
  IEEE/CVF Conference on Computer Vision and Pattern Recognition}, 2019, pp.
  7832--7841.

\bibitem{kr2019towards}
P.~KR, R.~Mukhopadhyay, J.~Philip, A.~Jha, V.~Namboodiri, and C.~Jawahar,
  ``Towards automatic face-to-face translation,'' in \emph{Proceedings of the
  27th ACM international conference on multimedia}, 2019, pp. 1428--1436.

\bibitem{vougioukas2020realistic}
K.~Vougioukas, S.~Petridis, and M.~Pantic, ``Realistic speech-driven facial
  animation with gans,'' \emph{International Journal of Computer Vision}, vol.
  128, no.~5, pp. 1398--1413, 2020.

\bibitem{wang2022progressive}
D.~Wang, Y.~Deng, Z.~Yin, H.-Y. Shum, and B.~Wang, ``Progressive disentangled
  representation learning for fine-grained controllable talking head
  synthesis,'' \emph{arXiv preprint arXiv:2211.14506}, 2022.

\bibitem{liu2023opt}
J.~Liu, X.~Wang, X.~Fu, Y.~Chai, C.~Yu, J.~Dai, and J.~Han, ``Opt: One-shot
  pose-controllable talking head generation,'' in \emph{ICASSP 2023-2023 IEEE
  International Conference on Acoustics, Speech and Signal Processing
  (ICASSP)}.\hskip 1em plus 0.5em minus 0.4em\relax IEEE, 2023, pp. 1--5.

\bibitem{zhou2020makelttalk}
Y.~Zhou, X.~Han, E.~Shechtman, J.~Echevarria, E.~Kalogerakis, and D.~Li,
  ``Makelttalk: speaker-aware talking-head animation,'' \emph{ACM Transactions
  on Graphics (TOG)}, vol.~39, no.~6, pp. 1--15, 2020.

\bibitem{wang2021audio2head}
S.~Wang, L.~Li, Y.~Ding, C.~Fan, and X.~Yu, ``Audio2head: Audio-driven one-shot
  talking-head generation with natural head motion,'' \emph{arXiv preprint
  arXiv:2107.09293}, 2021.

\bibitem{wang2022one}
S.~Wang, L.~Li, Y.~Ding, and X.~Yu, ``One-shot talking face generation from
  single-speaker audio-visual correlation learning,'' in \emph{Proceedings of
  the AAAI Conference on Artificial Intelligence}, vol.~36, no.~3, 2022, pp.
  2531--2539.

\bibitem{suwajanakorn2017synthesizing}
S.~Suwajanakorn, S.~M. Seitz, and I.~Kemelmacher-Shlizerman, ``Synthesizing
  obama: learning lip sync from audio,'' \emph{ACM Transactions on Graphics
  (ToG)}, vol.~36, no.~4, pp. 1--13, 2017.

\bibitem{fried2019text}
O.~Fried, A.~Tewari, M.~Zollh{\"o}fer, A.~Finkelstein, E.~Shechtman, D.~B.
  Goldman, K.~Genova, Z.~Jin, C.~Theobalt, and M.~Agrawala, ``Text-based
  editing of talking-head video,'' \emph{ACM Transactions on Graphics (TOG)},
  vol.~38, no.~4, pp. 1--14, 2019.

\bibitem{zhang2021facial}
C.~Zhang, Y.~Zhao, Y.~Huang, M.~Zeng, S.~Ni, M.~Budagavi, and X.~Guo, ``Facial:
  Synthesizing dynamic talking face with implicit attribute learning,'' in
  \emph{Proceedings of the IEEE/CVF international conference on computer
  vision}, 2021, pp. 3867--3876.

\bibitem{lahiri2021lipsync3d}
A.~Lahiri, V.~Kwatra, C.~Frueh, J.~Lewis, and C.~Bregler, ``Lipsync3d:
  Data-efficient learning of personalized 3d talking faces from video using
  pose and lighting normalization,'' in \emph{Proceedings of the IEEE/CVF
  conference on computer vision and pattern recognition}, 2021, pp. 2755--2764.

\bibitem{guo2021ad}
Y.~Guo, K.~Chen, S.~Liang, Y.-J. Liu, H.~Bao, and J.~Zhang, ``Ad-nerf: Audio
  driven neural radiance fields for talking head synthesis,'' in
  \emph{Proceedings of the IEEE/CVF International Conference on Computer
  Vision}, 2021, pp. 5784--5794.

\bibitem{mildenhall2021nerf}
B.~Mildenhall, P.~P. Srinivasan, M.~Tancik, J.~T. Barron, R.~Ramamoorthi, and
  R.~Ng, ``Nerf: Representing scenes as neural radiance fields for view
  synthesis,'' \emph{Communications of the ACM}, vol.~65, no.~1, pp. 99--106,
  2021.

\bibitem{chen2018lip}
L.~Chen, Z.~Li, R.~K. Maddox, Z.~Duan, and C.~Xu, ``Lip movements generation at
  a glance,'' in \emph{Proceedings of the European Conference on Computer
  Vision (ECCV)}, 2018, pp. 520--535.

\bibitem{chen2020talking}
L.~Chen, G.~Cui, C.~Liu, Z.~Li, Z.~Kou, Y.~Xu, and C.~Xu, ``Talking-head
  generation with rhythmic head motion,'' in \emph{European Conference on
  Computer Vision}.\hskip 1em plus 0.5em minus 0.4em\relax Springer, 2020, pp.
  35--51.

\bibitem{liang2022expressive}
B.~Liang, Y.~Pan, Z.~Guo, H.~Zhou, Z.~Hong, X.~Han, J.~Han, J.~Liu, E.~Ding,
  and J.~Wang, ``Expressive talking head generation with granular audio-visual
  control,'' in \emph{Proceedings of the IEEE/CVF Conference on Computer Vision
  and Pattern Recognition}, 2022, pp. 3387--3396.

\bibitem{yu2020multimodal}
L.~Yu, J.~Yu, M.~Li, and Q.~Ling, ``Multimodal inputs driven talking face
  generation with spatial--temporal dependency,'' \emph{IEEE Transactions on
  Circuits and Systems for Video Technology}, vol.~31, no.~1, pp. 203--216,
  2020.

\bibitem{song2022audio}
L.~Song, W.~Wu, C.~Fu, C.~C. Loy, and R.~He, ``Audio-driven dubbing for user
  generated contents via style-aware semi-parametric synthesis,'' \emph{IEEE
  Transactions on Circuits and Systems for Video Technology}, 2022.

\bibitem{siarohin2019first}
A.~Siarohin, S.~Lathuili{\`e}re, S.~Tulyakov, E.~Ricci, and N.~Sebe, ``First
  order motion model for image animation,'' \emph{Advances in Neural
  Information Processing Systems}, vol.~32, pp. 7137--7147, 2019.

\bibitem{ha2020marionette}
S.~Ha, M.~Kersner, B.~Kim, S.~Seo, and D.~Kim, ``Marionette: Few-shot face
  reenactment preserving identity of unseen targets,'' in \emph{Proceedings of
  the AAAI Conference on Artificial Intelligence}, vol.~34, no.~07, 2020, pp.
  10\,893--10\,900.

\bibitem{zhang2020freenet}
J.~Zhang, X.~Zeng, M.~Wang, Y.~Pan, L.~Liu, Y.~Liu, Y.~Ding, and C.~Fan,
  ``Freenet: Multi-identity face reenactment,'' in \emph{Proceedings of the
  IEEE/CVF Conference on Computer Vision and Pattern Recognition}, 2020, pp.
  5326--5335.

\bibitem{liu2021li}
J.~Liu, P.~Chen, T.~Liang, Z.~Li, C.~Yu, S.~Zou, J.~Dai, and J.~Han, ``Li-net:
  Large-pose identity-preserving face reenactment network,'' in \emph{2021 IEEE
  International Conference on Multimedia and Expo (ICME)}.\hskip 1em plus 0.5em
  minus 0.4em\relax IEEE, 2021, pp. 1--6.

\bibitem{ren2023hr}
Q.~Ren, Z.~Lu, H.~Wu, J.~Zhang, and Z.~Dong, ``Hr-net: a landmark based high
  realistic face reenactment network,'' \emph{IEEE Transactions on Circuits and
  Systems for Video Technology}, 2023.

\bibitem{bulat2017far}
A.~Bulat and G.~Tzimiropoulos, ``How far are we from solving the 2d \& 3d face
  alignment problem? (and a dataset of 230,000 3d facial landmarks),'' in
  \emph{International Conference on Computer Vision}, 2017.

\bibitem{zhang2017detecting}
K.~Zhang, Z.~Zhang, H.~Wang, Z.~Li, Y.~Qiao, and W.~Liu, ``Detecting faces
  using inside cascaded contextual cnn,'' in \emph{Proceedings of the IEEE
  International Conference on Computer Vision}, 2017, pp. 3171--3179.

\bibitem{pix2pix2017}
P.~Isola, J.-Y. Zhu, T.~Zhou, and A.~A. Efros, ``Image-to-image translation
  with conditional adversarial networks,'' \emph{CVPR}, 2017.

\bibitem{wang2018high}
T.-C. Wang, M.-Y. Liu, J.-Y. Zhu, A.~Tao, J.~Kautz, and B.~Catanzaro,
  ``High-resolution image synthesis and semantic manipulation with conditional
  gans,'' in \emph{Proceedings of the IEEE conference on computer vision and
  pattern recognition}, 2018, pp. 8798--8807.

\bibitem{thies2016face2face}
J.~Thies, M.~Zollhofer, M.~Stamminger, C.~Theobalt, and M.~Nie{\ss}ner,
  ``Face2face: Real-time face capture and reenactment of rgb videos,'' in
  \emph{Proceedings of the IEEE conference on computer vision and pattern
  recognition}, 2016, pp. 2387--2395.

\bibitem{shen2018faceid}
Y.~Shen, P.~Luo, J.~Yan, X.~Wang, and X.~Tang, ``Faceid-gan: Learning a
  symmetry three-player gan for identity-preserving face synthesis,'' in
  \emph{Proceedings of the IEEE conference on computer vision and pattern
  recognition}, 2018, pp. 821--830.

\bibitem{nagano2018pagan}
K.~Nagano, J.~Seo, J.~Xing, L.~Wei, Z.~Li, S.~Saito, A.~Agarwal, J.~Fursund,
  and H.~Li, ``pagan: real-time avatars using dynamic textures,'' \emph{ACM
  Transactions on Graphics (TOG)}, vol.~37, no.~6, pp. 1--12, 2018.

\bibitem{yao2020mesh}
G.~Yao, Y.~Yuan, T.~Shao, and K.~Zhou, ``Mesh guided one-shot face reenactment
  using graph convolutional networks,'' in \emph{Proceedings of the 28th ACM
  International Conference on Multimedia}, 2020, pp. 1773--1781.

\bibitem{ren2021pirenderer}
Y.~Ren, G.~Li, Y.~Chen, T.~H. Li, and S.~Liu, ``Pirenderer: Controllable
  portrait image generation via semantic neural rendering,'' in
  \emph{Proceedings of the IEEE/CVF International Conference on Computer
  Vision}, 2021, pp. 13\,759--13\,768.

\bibitem{peng2021unified}
B.~Peng, H.~Fan, W.~Wang, J.~Dong, and S.~Lyu, ``A unified framework for high
  fidelity face swap and expression reenactment,'' \emph{IEEE Transactions on
  Circuits and Systems for Video Technology}, vol.~32, no.~6, pp. 3673--3684,
  2021.

\bibitem{blanz1999morphable}
V.~Blanz and T.~Vetter, ``A morphable model for the synthesis of 3d faces,'' in
  \emph{Proceedings of the 26th annual conference on Computer graphics and
  interactive techniques}, 1999, pp. 187--194.

\bibitem{deng2019accurate}
Y.~Deng, J.~Yang, S.~Xu, D.~Chen, Y.~Jia, and X.~Tong, ``Accurate 3d face
  reconstruction with weakly-supervised learning: From single image to image
  set,'' in \emph{IEEE Computer Vision and Pattern Recognition Workshops},
  2019.

\bibitem{siarohin2019animating}
A.~Siarohin, S.~Lathuili{\`e}re, S.~Tulyakov, E.~Ricci, and N.~Sebe,
  ``Animating arbitrary objects via deep motion transfer,'' in
  \emph{Proceedings of the IEEE/CVF Conference on Computer Vision and Pattern
  Recognition}, 2019, pp. 2377--2386.

\bibitem{hong2022depth}
F.-T. Hong, L.~Zhang, L.~Shen, and D.~Xu, ``Depth-aware generative adversarial
  network for talking head video generation,'' in \emph{Proceedings of the
  IEEE/CVF Conference on Computer Vision and Pattern Recognition}, 2022, pp.
  3397--3406.

\bibitem{zhao2022thin}
J.~Zhao and H.~Zhang, ``Thin-plate spline motion model for image animation,''
  in \emph{Proceedings of the IEEE/CVF Conference on Computer Vision and
  Pattern Recognition}, 2022, pp. 3657--3666.

\bibitem{zakharov2019few}
E.~Zakharov, A.~Shysheya, E.~Burkov, and V.~Lempitsky, ``Few-shot adversarial
  learning of realistic neural talking head models,'' in \emph{Proceedings of
  the IEEE/CVF International Conference on Computer Vision}, 2019, pp.
  9459--9468.

\bibitem{burkov2020neural}
E.~Burkov, I.~Pasechnik, A.~Grigorev, and V.~Lempitsky, ``Neural head
  reenactment with latent pose descriptors,'' in \emph{Proceedings of the
  IEEE/CVF conference on computer vision and pattern recognition}, 2020, pp.
  13\,786--13\,795.

\bibitem{wang2020imaginator}
Y.~Wang, P.~Bilinski, F.~Bremond, and A.~Dantcheva, ``Imaginator: Conditional
  spatio-temporal gan for video generation,'' in \emph{Proceedings of the
  IEEE/CVF Winter Conference on Applications of Computer Vision}, 2020, pp.
  1160--1169.

\bibitem{zeng2020realistic}
X.~Zeng, Y.~Pan, M.~Wang, J.~Zhang, and Y.~Liu, ``Realistic face reenactment
  via self-supervised disentangling of identity and pose,'' in
  \emph{Proceedings of the AAAI Conference on Artificial Intelligence},
  vol.~34, no.~07, 2020, pp. 12\,757--12\,764.

\bibitem{karras2019style}
T.~Karras, S.~Laine, and T.~Aila, ``A style-based generator architecture for
  generative adversarial networks,'' in \emph{Proceedings of the IEEE/CVF
  conference on computer vision and pattern recognition}, 2019, pp. 4401--4410.

\bibitem{karras2020analyzing}
T.~Karras, S.~Laine, M.~Aittala, J.~Hellsten, J.~Lehtinen, and T.~Aila,
  ``Analyzing and improving the image quality of stylegan,'' in
  \emph{Proceedings of the IEEE/CVF conference on computer vision and pattern
  recognition}, 2020, pp. 8110--8119.

\bibitem{chen2023compact}
B.~Chen, Z.~Wang, B.~Li, S.~Wang, and Y.~Ye, ``Compact temporal trajectory
  representation for talking face video compression,'' \emph{IEEE Transactions
  on Circuits and Systems for Video Technology}, 2023.

\bibitem{weston2014memory}
J.~Weston, S.~Chopra, and A.~Bordes, ``Memory networks,'' \emph{arXiv preprint
  arXiv:1410.3916}, 2014.

\bibitem{miller2016key}
A.~Miller, A.~Fisch, J.~Dodge, A.-H. Karimi, A.~Bordes, and J.~Weston,
  ``Key-value memory networks for directly reading documents,'' \emph{arXiv
  preprint arXiv:1606.03126}, 2016.

\bibitem{yi2020audio}
R.~Yi, Z.~Ye, J.~Zhang, H.~Bao, and Y.-J. Liu, ``Audio-driven talking face
  video generation with learning-based personalized head pose,'' \emph{arXiv
  preprint arXiv:2002.10137}, 2020.

\bibitem{jiang2022text}
S.~Jiang, Y.~Shi, and K.~Cheng, ``Text-to-face generation via multi-modal
  attention memory network with fine-grained feedback,'' in \emph{2022 5th
  International Conference on Pattern Recognition and Artificial Intelligence
  (PRAI)}.\hskip 1em plus 0.5em minus 0.4em\relax IEEE, 2022, pp. 940--947.

\bibitem{paysan20093d}
P.~Paysan, R.~Knothe, B.~Amberg, S.~Romdhani, and T.~Vetter, ``A 3d face model
  for pose and illumination invariant face recognition,'' in \emph{2009 sixth
  IEEE international conference on advanced video and signal based
  surveillance}.\hskip 1em plus 0.5em minus 0.4em\relax Ieee, 2009, pp.
  296--301.

\bibitem{isola2017image}
P.~Isola, J.-Y. Zhu, T.~Zhou, and A.~A. Efros, ``Image-to-image translation
  with conditional adversarial networks,'' in \emph{Proceedings of the IEEE
  conference on computer vision and pattern recognition}, 2017, pp. 1125--1134.

\bibitem{huang2017arbitrary}
X.~Huang and S.~Belongie, ``Arbitrary style transfer in real-time with adaptive
  instance normalization,'' in \emph{Proceedings of the IEEE international
  conference on computer vision}, 2017, pp. 1501--1510.

\bibitem{simonyan2014very}
K.~Simonyan and A.~Zisserman, ``Very deep convolutional networks for
  large-scale image recognition,'' \emph{arXiv preprint arXiv:1409.1556}, 2014.

\bibitem{chung2016lip}
J.~S. Chung and A.~Zisserman, ``Lip reading in the wild,'' in \emph{Asian
  conference on computer vision}.\hskip 1em plus 0.5em minus 0.4em\relax
  Springer, 2016, pp. 87--103.

\bibitem{chung2018voxceleb2}
J.~S. Chung, A.~Nagrani, and A.~Zisserman, ``Voxceleb2: Deep speaker
  recognition,'' \emph{arXiv preprint arXiv:1806.05622}, 2018.

\bibitem{wang2004image}
Z.~Wang, A.~C. Bovik, H.~R. Sheikh, and E.~P. Simoncelli, ``Image quality
  assessment: from error visibility to structural similarity,'' \emph{IEEE
  transactions on image processing}, vol.~13, no.~4, pp. 600--612, 2004.

\bibitem{narvekar2009no}
N.~D. Narvekar and L.~J. Karam, ``A no-reference perceptual image sharpness
  metric based on a cumulative probability of blur detection,'' in \emph{2009
  International Workshop on Quality of Multimedia Experience}.\hskip 1em plus
  0.5em minus 0.4em\relax IEEE, 2009, pp. 87--91.

\bibitem{chung2016out}
J.~S. Chung and A.~Zisserman, ``Out of time: automated lip sync in the wild,''
  in \emph{Asian conference on computer vision}.\hskip 1em plus 0.5em minus
  0.4em\relax Springer, 2016, pp. 251--263.

\bibitem{ruiz2018fine}
N.~Ruiz, E.~Chong, and J.~M. Rehg, ``Fine-grained head pose estimation without
  keypoints,'' in \emph{Proceedings of the IEEE conference on computer vision
  and pattern recognition workshops}, 2018, pp. 2074--2083.

\end{thebibliography}

\end{document}